\documentclass[letterpaper]{article} 
\usepackage[utf8]{inputenc} 
\usepackage{aaai2026}  
\usepackage{times}  
\usepackage{helvet}  
\usepackage{courier}  
\usepackage[hyphens]{url}  
\usepackage{graphicx} 
\usepackage[table]{xcolor} 
\usepackage{tabularx,booktabs,array}
\newcolumntype{B}{>{\hsize=2\hsize\raggedright\arraybackslash}X}   
\newcolumntype{S}{>{\hsize=.6\hsize\centering\arraybackslash}X}  
\newcolumntype{L}[1]{>{\raggedright\arraybackslash}p{#1}}
\newcolumntype{C}{>{\centering\arraybackslash}X}
\newcolumntype{Y}{>{\centering\arraybackslash}X}

\usepackage{natbib}  
\usepackage{caption} 
\usepackage{algorithm}
\usepackage{algorithmic}

\usepackage{newfloat}
\usepackage{listings}
\DeclareCaptionStyle{ruled}{labelfont=normalfont,labelsep=colon,strut=off} 
\floatstyle{ruled}
\newfloat{listing}{tb}{lst}{}
\floatname{listing}{Listing}
\usepackage{amsmath}

\lstdefinelanguage{json}{
  basicstyle=\ttfamily\footnotesize,
  upquote=true,
  showstringspaces=false,
  breaklines=true,
  frame=single,
  literate=
   *{0}{{{\color{black}0}}}{1}
    {1}{{{\color{black}1}}}{1}
    {2}{{{\color{black}2}}}{1}
    {3}{{{\color{black}3}}}{1}
    {4}{{{\color{black}4}}}{1}
    {5}{{{\color{black}5}}}{1}
    {6}{{{\color{black}6}}}{1}
    {7}{{{\color{black}7}}}{1}
    {8}{{{\color{black}8}}}{1}
    {9}{{{\color{black}9}}}{1}
    {:}{{{\color{black}{:}}}}{1}
    {,}{{{\color{black}{,}}}}{1}
    {\{}{{{\color{black}{\{}}}}{1}
    {\}}{{{\color{black}{\}}}}}{1}
    {[}{{{\color{black}{[}}}}{1}
    {]}{{{\color{black}{]}}}}{1}
}
\lstdefinestyle{jsonlight}{
  language=json,
  basicstyle=\ttfamily\footnotesize,
  numbers=left,
  numbersep=6pt,
  xleftmargin=1em,
  frame=single,
  breaklines=true,
}

\title{Executable Governance for AI: Translating Policies into Rules Using LLMs}

\author{ Gautam Varma Datla\textsuperscript{\rm 1}, Anudeep Vurity\textsuperscript{\rm 2}, Tejaswani Dash\textsuperscript{\rm 3}, Tazeem Ahmad\textsuperscript{\rm 4},\\ Mohd Adnan\textsuperscript{\rm 5}, Saima Rafi\textsuperscript{\rm 6} }

\affiliations{
\textsuperscript{\rm 1}New Jersey Institute of Technology, Newark, USA\\
\textsuperscript{\rm 2,3}George Mason University, Fairfax, USA\\
\textsuperscript{\rm 4}School of Mathematics, Physics and Computing, University of Southern Queensland, Australia\\
\textsuperscript{\rm 5}University of Aveiro, Aveiro, Portugal\\
\textsuperscript{\rm 6}Edinburgh Napier University, Edinburgh, Scotland, UK\\[2pt]
\textsuperscript{\rm 1}gvd6@njit.edu,
\textsuperscript{\rm 2}avurity@gmu.edu,
\textsuperscript{\rm 3}tdash@gmu.edu,
\textsuperscript{\rm 4}Tazeem.Ahmad@unisq.edu.au,
\textsuperscript{\rm 5}m.adnan1821@gmail.com,
\textsuperscript{\rm 6}s.rafi@napier.ac.uk
}

\usepackage[T1]{fontenc}
\usepackage{textcomp}

\begin{document}

\maketitle

\begin{abstract}
AI policy guidance is predominantly written as prose, which practitioners must first convert into executable rules before frameworks can evaluate or enforce them. This manual step is slow, error-prone, difficult to scale, and often delays the use of safeguards in real-world deployments. To address this gap, we present Policy$\rightarrow$Tests (P2T), a framework that converts natural-language policy documents into normalized, machine-readable rules. The framework comprises a pipeline and a compact domain-specific language (DSL) that encodes hazards, scope, conditions, exceptions, and required evidence, yielding a canonical representation of extracted rules. To test the framework beyond a single policy, we apply it across general frameworks, sector guidance, and enterprise standards, extracting obligation-bearing clauses and converting them into executable rules. These AI-generated rules closely match strong human baselines on span- and rule-level metrics, with robust inter-annotator agreement on the gold set. To evaluate downstream behavioral and safety impact, we add HIPAA-derived safeguards to a generative agent and compare it with an otherwise identical agent without guardrails. An LLM-based judge, aligned with gold-standard criteria, measures violation rates and robustness to obfuscated and compositional prompts. Detailed results are provided in the appendix. We release the codebase, DSL, prompts, and rule sets as open-source resources to enable reproducible evaluation.
\end{abstract}

\begin{links}
    \link{Code}{https://github.com/gautamvarmadatla/Policy-Tests-P2T-for-operationalizing-AI-governance}
\end{links}

\section{Introduction}

As artificial intelligence (AI) permeates critical sectors, ensuring its \textit{responsible use} has become imperative \cite{cheng2025pii,shen2024jailbreak,carlini2021extracting,wei2024cotaeval}. In response, governments and industry bodies have introduced governance frameworks. For example, the European Union AI Act sets obligations for high risk uses \cite{EU_2024_AI_Act}, and the NIST AI Risk Management Framework provides a voluntary guide focused on trustworthiness \cite{NIST_2023_AI_RMF_1_0}. Building on these efforts, additional standards and principles promote responsible practice, including the OECD AI Principles \cite{OECD_2019_AI_Principles} and ISO IEC 42001 for AI management systems \cite{ISOIEC_42001_2023}. However, these instruments are intentionally nonprescriptive. The AI RMF explicitly states that it is intended to be voluntary, rights preserving, and use case agnostic, and the companion Playbook offers suggestions that organizations may adopt as needed, rather than mandated tests \cite{NIST_2023_Playbook}. Efforts like Singapore’s AI Verify show how process checks and technical evaluations can be packaged into a single toolkit, yet teams still spend expert hours turning broad guidance into the concrete checks that a specific system will pass or fail in a repeatable way \cite{PDPC2021AIVerifyMVP}. From here, the gap unfolds in practice, as organizations must invent concrete, verifiable procedures to demonstrate compliance in deployed systems, and this shortfall can erode public trust in high-stakes settings.

Our work targets this gap by asking whether policy text can be turned into checks that fit naturally into an engineering workflow. We first ask whether an automated pipeline can extract policy rules with span level and field level quality that stands up to careful human work. We then ask whether rules derived from policy actually reduce observed violations when they are used to evaluate the behavior of AI agents under a consistent runtime. Finally, we examine which controls matter most for the quality and robustness of extracted rules, including schema guarded decoding, targeted repairs, evidence gating, semantic deduplication, SMT conflict checks, and counterfactual flips.

\begin{figure*}[t]
  \centering
  \includegraphics[width=\textwidth]{figure1.pdf}
  \caption{P2T overview. The pipeline reads policy documents and returns executable atomic rules. It does so by iteratively extracting and refining rules with LLMs and deterministic checks, including clause mining, evidence gating, and SMT (Satisfiability Modulo Theories) validation.}
  \label{fig:pipeline}
\end{figure*}

To address these questions we present a Policy$\rightarrow$Tests (P2T) framework that converts policy documents into executable rules without breaking stride. The pipeline ingests policy documents and finds spans that are likely to contain obligations or prohibitions, then performs schema guarded LLM extraction supported by deterministic checks and an optional LLM judge so that rules are both structured and scrutinized. When outputs fall short the system applies minimal edit repairs that preserve meaning, removes near duplicates through semantic clustering so that reviewers see the essential set, and flags contradictions through logical checks that help analysts resolve conflicts early. A compact machine readable schema captures hazards, scope, conditions, exceptions, and required evidence, which lets natural language clauses compile to checklists and scenario based rules that engineers can run. We evaluate P2T through human baselines, ablations over pipeline controls, and agreement audits comparing AI-generated rules with adjudicated human extractions for each document. Because there are no established benchmarks for policy-to-rule translation, we focus on internal consistency, human agreement, and controlled ablations rather than direct model-to-model comparisons.
We release the codebase, prompts, DSL, and public policy derived rule sets with testability flags to enable independent replication and reuse. Our goal and contribution is a clear path from principle to proof, where teams can routinely test for compliance, surface issues early in development, and sustain responsible deployment in settings where failures carry real consequences.

\section{Background \& Related Work}
In recent years, the community has pursued three main routes for operationalizing responsible AI. Training-time alignment, including RLHF and Constitutional AI, improves baseline behavior by internalizing a fixed set of principles \cite{ouyang2022instructgpt, bai2022constitutional}. These methods are valuable, but they adapt poorly to evolving or domain-specific policies because policy updates require new data, new fine-tuning, and there is no straightforward way to verify compliance with each concrete obligation. Runtime guardrail frameworks such as NeMo Guardrails and Guardrails AI offer a practical safety layer across models by enforcing developer-authored rules in a DSL \cite{ nemo_guardrails_overview, guardrails_ai_github}. Their effectiveness, however, is bounded by manual authoring and maintenance, and coverage remains limited to the rules practitioners remember to encode. In parallel, evaluation toolkits like OpenAI Evals and PromptFoo facilitate scripted assessments of model behavior, yet they presuppose that tests already exist, leaving the translation from policy prose to executable checks as a largely manual exercise \cite{openai_evals, promptfoo_docs}.

\begin{table*}[h]
\centering
\renewcommand{\arraystretch}{1.1} 
\setlength{\tabcolsep}{7pt}       
\fontsize{9}{10}\selectfont      
\begin{minipage}{0.83\textwidth}   
\centering
\begin{tabularx}{\textwidth}{@{}l c c c c c@{}}
\toprule
\textbf{Document} & 
\textbf{Cand.} & 
\textbf{Ext.} & 
\textbf{Uniq.} & 
\textbf{Test.\ (\%)} & 
\textbf{Ex.\ (IO)} \\
\midrule
EU AI Act (Reg.\ (EU) 2024/1689, Arts.\ 8--15) & 21 & 63 & 51 & 33.3 & 0 \\
NIST AI RMF (Profiles:\ MAP 1.1--5.3;\ MEASURE 1.1--2.7) & 60 & 122 & 117 & 43.6 & 7 \\
HIPAA Privacy Rule (45 C.F.R. Part 164 Subpart E) & 56 & 94 & 77 & 85.7 & 31 \\
Microsoft Responsible AI Standard v2 & 88 & 196 & 140 & 75.0 & 7 \\
\midrule
\textbf{Total} & 276 & 522 & 427 & 58.8 & 50 \\
\bottomrule
\end{tabularx}
\caption{Corpus summary and pipeline. Abbreviations: Cand.\ = Candidate spans, Ext.\ = Extracted rules, Uniq.\ = Unique rules after de-duplication, Test.\ = Testable rules after de-duplication, Ex.\ (I/O) = Testable rules with input/output validation as evidence signals.}
\label{tab:corpus-summary-wide}
\end{minipage}
\end{table*}

Adjacent efforts illuminate the gap without closing it. Policy-as-prompt approaches pass natural-language policies to models as classifiers or filters, but they still rely on careful hand-crafted prompts to reflect real clauses \cite{palla2025policyasprompt}. Rules-as-code platforms and policy engines such as OpenFisca, DMN, or OPA/Rego execute formal rules at scale, yet they begin only after experts have already converted text into logic \cite{openfisca_docs, omg_dmn_overview, opa_docs_intro}. Knowledge-graph and Graph-RAG pipelines help organize regulations and ground answers in source passages, but they emphasize retrieval and question answering rather than producing a portable, machine-readable rule corpus \cite{ibm_graphrag_tutorial}. Benchmarks including COMPL-AI, SG-Bench, TAU-Bench, and recent agent testbeds reveal policy failures under controlled scenarios, but they evaluate against predetermined criteria rather than deriving those criteria from governance documents \cite{complai2024,sgbench2024,toloka_taubench_policy_agents}. In response to these limitations, we provide the missing bridge from policy documents to enforcement and testing frameworks.  We automatically extract structured, machine-readable rules from policy text. The extracted rules record provenance, scope, and hazards; encode conditions and exceptions; specify concrete requirements and acceptable evidence; annotate severity and testability; and include illustrative benign and adversarial examples. They serve as reusable artifacts consumed by downstream enforcement and evaluation systems, rather than tests themselves. A single extracted rule can be rendered as a NeMo Colang snippet for runtime control, configured as a Guardrails validator, compiled into OPA/Rego for policy engines, or transformed into Evals-style prompts for batch assessment. The pipeline can be re-run on updated documents to track policy evolution without retraining. Moreover, it incorporates validation and repair procedures, including schema checks, LLM judging, targeted repair, and optional SMT consistency analysis, which improve fidelity to source clauses and reduce contradictions. Collectively, these properties position our approach as the missing policy-to-rule layer that connects policy specification to enforcement and evaluation, enabling continuous, auditable compliance.

\newcommand{\tablesize}{\scriptsize}   
\newcommand{\tablecolsep}{1.6pt}       
\renewcommand{\arraystretch}{1.02}     

\section{Pipeline Overview}
\label{sec:pipeline-overview}
We present a modular, provenance-preserving pipeline that converts policy text into reusable, machine-readable rules for enforcement and evaluation. Deterministic modules handle ingestion, clause filtering, schema validation, vocabulary normalization, de-duplication, and consistency checks to ensure stability and reproducibility, as seen in Fig.\ref{fig:pipeline}. LLMs are used only where semantic interpretation is essential: to generate atomic rules under a strict JSON schema, repair invalid rules, tag testability, and synthesize rule-specific examples. 

To make stages interoperable, we define a compact JSON DSL that fixes the rule schema; each stage then fills in or validates parts of this schema so the record is progressively completed. The DSL encodes each clause as a JSON rule with fixed fields. \texttt{scope} uses predefined enums for \texttt{actor}, \texttt{data\_domain}, and \texttt{context}. \texttt{hazard}, \texttt{conditions}, \texttt{exceptions}, and \texttt{requirement} are natural-language strings, with arrays when multiple items occur. \texttt{testability} carries a boolean flag, a short rationale, and \texttt{evidence\_signals} from a closed set (\texttt{io\_check}, \texttt{log\_check}, \texttt{config\_check}, \texttt{ci\_gate}, \texttt{data\_check}, \texttt{repo\_check}, \texttt{access\_check}, \texttt{attest\_check}). \texttt{evidence} is populated only when the clause explicitly names an artifact; otherwise it remains empty. Each rule records provenance (\texttt{doc}, \texttt{citation}, \texttt{span\_id}). The full DSL schema appears in the Appendix. We next walk through the pipeline stages, from ingestion to example generation.

\medskip
\noindent\textbf{Step 1: Ingestion and chunking.}
This step converts raw policy text into clean, addressable units that the pipeline can reason about. It removes boilerplate, preserves section context, and splits content into spans such as sentences, captions, figure summaries, and tables. The result is a consistent stream of spans, each with text, location context, and a stable identifier, creating a canonical evidence feed for clause mining and end-to-end traceability across all downstream stages. 

We primarily evaluated three chunking strategies for this step: single-sentence spans, sliding context windows that attach neighboring sentences to each target sentence, and paragraph-level spans treated as the atomic unit. For policy documents, paragraph-level spans worked best because they provide enough local context for grounded obligations without overwhelming the model. Sliding windows produced reasonable results but increased duplicates due to overlapping windows that encouraged the model to restate similar rules. Single-sentence spans minimized duplication but led to under-specified rules when clauses relied on surrounding qualifiers. Based on these trade-offs, we default to paragraph-level spans while retaining section headers and citations for provenance.

\medskip
\noindent\textbf{Step 2: Clause mining (Optional).}
This step narrows spans to sentences likely to encode enforceable policy by detecting deontic markers, exception cues, actor mentions, quantitative or temporal indicators, and cross-references. It downweights definitions and boilerplate, removes duplicates, and outputs candidate clauses with tags and provenance for structured extraction. The tags specify clause type (obligation, prohibition, exception, exemption, definition, or other) and extract deadlines, thresholds, and cross-references. These deterministic filters reduce downstream LLM cost but may lower recall when obligations lack explicit cues or span multiple bullets. Tailoring heuristics to an organization’s writing style improves recall while retaining the efficiency benefits of early filtering.

\begin{table*}[h]
\centering
\renewcommand{\arraystretch}{1.1} 
\setlength{\tabcolsep}{6pt}       
\fontsize{8}{10}\selectfont
\begin{minipage}{\textwidth}  
\centering
\begin{tabularx}{\textwidth}{@{}l *{8}{c}@{}}
\toprule
\textbf{Document} &
\shortstack[c]{\textbf{No. Gold}\\\textbf{Span}}&
\textbf{Pred} &
\textbf{Cov.}{↑} &
\textbf{TestAcc}{↑} &
\textbf{Span F$_1$}{↑}&
\shortstack[c]{\textbf{Span}\\\textbf{AUPRC ↑}} &
\shortstack[c]{\textbf{SE}\\\textbf{Similarity ↑}} &
\shortstack[c]{\textbf{Ev}\\\textbf{Similarity ↑}} \\
\midrule
\cite{EU_2024_AI_Act} &
20 & 11.00 & \(0.55 \pm 0.20\) & \(0.55 \pm 0.20\) & \(0.79 \pm 0.14\) & \(0.73 \pm 0.16\) & \(0.24 \pm 0.09\) & \(0.24 \pm 0.10\) \\
\cite{NIST_2023_AI_RMF_1_0} &
33 & 24.50 & \(0.71 \pm 0.13\) & \(0.71 \pm 0.13\) & \(0.83 \pm 0.11\) & \(0.88 \pm 0.11\) & \(0.26 \pm 0.05\) & \(0.30 \pm 0.07\) \\
\cite{hipaa} &
43 & 34.50 & \(0.76 \pm 0.11\) & \(0.76 \pm 0.11\) & \(0.75 \pm 0.08\) & \(0.70 \pm 0.13\) & \(0.26 \pm 0.05\) & \(0.32 \pm 0.06\) \\
\cite{microsoft} &
71 & 54.50 & \(0.75 \pm 0.08\) & \(0.75 \pm 0.08\) & \(0.84 \pm 0.06\) & \(0.91 \pm 0.06\) & \(0.30 \pm 0.04\) & \(0.36 \pm 0.06\) \\
\bottomrule
\end{tabularx}
\caption{Evaluation by document with span-level and rule-level metrics. 
Abbreviations: Pred = predicted spans, 
Cov. = coverage, TestAcc = testability accuracy, 
Span F$_1$ = span-level F1, Span AUPRC = span-level area under the precision–recall curve, SE Similarity = structured extraction slot-level similarity, 
Ev Similarity = evidence-signal field similarity. 
↑ indicates higher values are better. 
All values except Gold span represent means across decoding seeds, with “±” indicating 95\% bootstrap confidence intervals.}
\label{tab:eval-by-doc}
\end{minipage}
  \end{table*}

\medskip
\textbf{Step 3: Structured extraction.} This is the main step where the pipeline produces rules that systems can actually use. An LLM performs the initial extraction, guided by few-shot examples and a strict DSL-compliant JSON schema, to emit atomic, machine-readable rules. After structural validation and scope normalization, we invoke an LLM-based judge to flag missing hazards, empty scope, unverifiable evidence, or conflicts between requirements and exceptions. When it flags issues, a repair LLM applies minimal, provenance-preserving edits. Judge outputs were not double annotated; we validated their utility via the observed gains with Judge+Repair in Table 3. After this, we optionally apply three checks: (1) an evidence gate that enforces verifiability by requiring appropriate evidence fields and, if configured, restricting sources to trusted domains; (2) a Satisfiability Modulo Theories (SMT) consistency check that encodes rules as logical constraints and uses a solver to detect contradictions where overlapping scopes would require and forbid the same predicate; and (3) counterfactual probing that generates small paraphrases of the source clause to test polarity sensitivity and expose fragile extractions or overfitting. The output is an aggregated rules file and a per-clause trace that record accepted rules, confidence, and any issues, ready for enforcement and evaluation. All prompts used in this step are available in our repository.

\medskip
\noindent\textbf{Step 4: De-duplication.}
This step consolidates repeated or paraphrased rules so the resulting rule set reflects unique obligations without losing traceability. We first apply structural de-duplication using a canonical signature over scope, hazard, conditions, exceptions, requirements, and severity, aggregating all contributing spans. We then run a semantic pass that embeds each rule and merges high-similarity pairs within sensible blocks such as the same document and scope. For example, from the EU AI Act, Regulation (EU) 2024/1689, Article 10(5)(a) and 10(5)(f), we extract two rules from one clause: providers must “verify and document that other data, including synthetic or anonymised, would not suffice for bias detection” and providers must “ensure the records of processing activities state why special-category data were necessary and why alternatives would not work.” Although their evidence slots differ, both express one obligation, so the semantic pass merged them into a single canonical rule. 

\medskip
\noindent\textbf{Step 5: Testability tagging (Optional).}
Here we assess whether each rule can be operationally verified. An LLM reviews the rule and, using a fixed rubric, determines if an objective pass–fail oracle exists and which evidence channels apply (e.g., I/O inspection, logs, configuration, CI artifacts, data, or repository state). The output augments each rule with \texttt{is\_testable}, a short rationale, and suggested evidence signals. This optional but recommended step produces rules that can be tested, with defined evidence channels simplifying check implementation in downstream frameworks.

\medskip
\noindent\textbf{Step 6: Example generation (Optional).}
For rules that are marked testable and include an I/O evidence signal, an LLM generates small sets of realistic prompts: benign cases that should pass the rule and adversarial cases that should provoke a violation if the rule is enforced. Each set is tailored to the rule’s scope, hazard, conditions, exceptions, and severity, and follows strict JSON output. This creates immediate, organization-specific test inputs for black-box evaluation and regression, enriching each rule with reproducible examples that downstream enforcement and assessment harnesses can run as-is.

\section{Corpus, Baselines \& Evaluation }
\subsection{Corpus}

This keeps the schema grounded in the obligations practitioners actually face and improves coverage and transfer across domains. We also select sources that clearly state obligations or prohibitions, articulate conditions and exceptions, and vary in drafting style, so evaluation reflects legal, technical, and operational language. Concretely, we mine selected titles from the EU AI Act (Reg.\ (EU) 2024/1689) Articles 8--15 \cite{EU_2024_AI_Act}; the NIST AI RMF Profiles (MAP 1.1--5.3 and MEASURE 1.1--2.7) \cite{NIST_2023_AI_RMF_1_0}; Microsoft Responsible AI Standard v2 \cite{microsoft}; and the HIPAA Privacy Rule, 45 C.F.R.\ Part 164 \cite{hipaa}. Each document passes through the pipeline to produce spans, then candidate clauses, with light neighboring context retained to aid interpretation. Human reviewers then convert the sampled clauses into atomic rules. For transparency, we report per-document counts at each pipeline stage as shown in Table~\ref{tab:corpus-summary-wide}, where generated examples are produced only for rules tagged as testable with I/O-check evidence signals. All public texts are cited and used under their licenses, enterprise policies are taken from public postings or short excerpts under fair use, and no personal data is processed.

\begin{table*}[t]
\centering
\begin{minipage}{0.94\textwidth}
\centering
\small
\resizebox{\textwidth}{!}{%
\setlength{\tabcolsep}{2pt}
\fontsize{8}{12}\selectfont
\begin{tabularx}{\textwidth}{
  l
  c c c
  c c
  c c c c c c c
}
\toprule
\textbf{Variant} &
\shortstack[c]{\textbf{Coverage}} &
\shortstack[c]{\textbf{Span F$_1$}} &
\shortstack[c]{\textbf{SE}\\\textbf{slot similarity}} &
\shortstack[c]{\textbf{DupIdx}$\downarrow$} &
\shortstack[c]{\textbf{Actor}\\\textbf{similarity}} &
\shortstack[c]{\textbf{Requirements}\\\textbf{similarity}} &
\shortstack[c]{\textbf{Conditions}\\\textbf{similarity}} &
\shortstack[c]{\textbf{Exceptions}\\\textbf{similarity}} &
\shortstack[c]{\textbf{Evidence signal}\\\textbf{similarity}} \\
\midrule
GPT-5-mini ( Few-Shot ) &
0.9296 & 0.9714 & 0.3641 & 0.0000 &
0.5352  & 0.5388 & 0.4447 & 0.9127 & 0.5155 \\
+ Judge + Repair  &
0.9437 & 0.9784 & 0.3886 & 0.0000 &
0.6620  & 0.5633 & 0.4316 & 0.9280 & 0.5904 \\
+ Judge + Repair + Dedup  &
0.8873 & 0.9640 & 0.3624 & 0.1525 &
0.5775  & 0.5009 & 0.4254 & 0.8592 & 0.5378 \\
\arrayrulecolor{black!20}\midrule
\arrayrulecolor{black}
GPT-4o-mini ( Few-Shot ) &
0.8875 & 0.8342 & 0.2528 & 0.0000  &
0.6324 & 0.3951 & 0.3009 & 0.8718 & 0.3014 \\
+ Judge + Repair  &
0.8028 & 0.8824 & 0.2944 & 0.0000 &
0.6197 & 0.2693  & 0.2921  & 0.7887  & 0.3357\\
+ Judge + Repair + Dedup &
0.8169 & 0.8759 & 0.2949 & 0.2039 &
0.6338 & 0.2760 & 0.2772 & 0.7887 & 0.3521 \\
\bottomrule
\end{tabularx}%
}
\caption{Ablation results. DupIdx = duplicates removed divided by the sum of rules kept after de-duplication and duplicates removed. Span F$_1$ = precision recall F$_1$ on span matches. SE slot similarity = mean of span averages of field similarities. Actor Requirements Conditions Exceptions Evidence signal = per field similarities; actor uses minimum one item overlap, others use LLM similarity in 0 to 1. Ablations begin from a simple LLM (few-shot) extraction baseline under our DSL schema, then add Judge+Repair and Dedup in sequence.}
\label{tab:ablation-slot-final}
\end{minipage}
\end{table*}

\subsection{Human-Annotated Gold Set and Protocol}
We build a compact, human-annotated gold set that reflects the end-to-end policy-to-rule pipeline described above. Annotators first decide, at the span-level, whether a span expresses one or more atomic rules. For positive spans, they extract the atomic rules and complete fields for hazard; scope (actor, data domain, context); conditions; exceptions; requirements; and evidence. They also judge whether each rule is operationally testable and select the evidence channels that could verify it. Each span is annotated independently by two raters and disagreements are adjudicated by a senior reviewer. We report agreement at the span level and the field level, using \textbf{Cohen’s $\kappa$} \cite{cohen1960} for categorical labels and \textbf{Krippendorff’s $\alpha$} \cite{krippendorff2018} for multi-label sets, with 95\% confidence intervals. Here, span level refers to the sentence window used to decide whether it contains one or more atomic rules, and field level is computed per extracted rule slot. Quality control proceeds in three steps that connect to the main task. First, a 30-item calibration set is completed to align raters; any low-agreement patterns trigger targeted rubric edits before full-scale annotation. Second, to track drift over time, we seed 5\% of assignments with previously adjudicated hidden-gold items and monitor agreement on these checks. Third, we stratify sampling by source and domain so the resulting set remains balanced and informative throughout the study. Across five source-specific gold sets ($n_{\text{rules}}{=}427$; per-doc $n{=}\{51,117,42,77,140\}$), the macro-average (unweighted over docs) was: span $\kappa = 0.83 \pm 0.03$, testable $\kappa = 0.76 \pm 0.04$, scope-actors $\alpha = 0.63 \pm 0.05$ and hazard $\kappa = 0.64 \pm 0.05$.

\subsection{Evaluation Setup, Metrics, and Ablations}

Two seeds are run for every document and they differ only in sampling so any variation reflects generation randomness. For each document we evaluate each seed against its human gold set. To measure quality of structured extraction we compute field similarity for each field. Actor, domain and context use a minimum one item overlap between the gold list and the predicted list which yields a score of one for a hit and zero otherwise. Hazard, conditions, exceptions, requirements and evidence signals use an LLM based similarity score in the range zero to one with a token-based Jaccard fallback when the corresponding filed is unavailable. For each span we average the field similarity values to obtain a span macro similarity. We then average span macro similarity across all spans to report \textit{SE slot similarity}. \textit{Coverage} is the fraction of gold spans that have at least one predicted rule that matches the span by exact identifier or by citation tail match. \textit{Testable accuracy} compares the predicted \texttt{is\_testable} label with the gold label when present. We also report \textit{evidence signal similarity} by averaging the evidence signals field scores over spans. For span detection we treat unique predicted spans extracted from rules as detections of gold spans. A predicted span is a true positive if it matches any gold span by exact identifier or by citation tail match. Otherwise it is a false positive. A gold span without a matching prediction is a false negative. For span-level metrics, we report \textit{F1-score} and \textit{Area under the precision recall curve ( AUPRC)}. Results are reported in Table \ref{tab:eval-by-doc}. 

\textbf{Ablations}. We primarily vary two components to assess their impact on rule quality. First the extraction model using GPT 5 mini and GPT 4o mini to test robustness across architectures, disentangle pipeline effects from model capacity and show that gains are not tied to a single configuration or provider.  Second the safeguards are applied in stages as LLM $\rightarrow$ LLM + judge + repair $\rightarrow$ LLM + judge + repair + deduplication. Results appear in Table \ref{tab:ablation-slot-final}: Adding Judge + Repair improves coverage, span-level F1, and SE slot similarity across both GPT-5-mini and GPT-4o-mini, while also boosting alignment on actor roles and evidence signals. Adding semantic de-duplication further enhances deduplication and actor consistency. However, it slightly reduces coverage and field-level similarity, indicating a trade-off between redundancy reduction and recall.

\section{Conclusion}
Policy$\rightarrow$Tests translates policy prose into executable, machine-readable rules through a validated pipeline and a compact DSL, and it generalizes across heterogeneous documents. The resulting rules align closely with human baselines and improve downstream behavior. However, limitations remain: ambiguous or context-dependent language may yield incomplete or incorrect rules, and a compact DSL might not capture all nuances (e.g., temporal or probabilistic constraints). Furthermore, relying on LLMs introduces risks, because models can hallucinate or generate plausible but incorrect rules, so rigorous validation is required. As prior work notes, human oversight remains essential in high-stakes rule extraction~\cite{kennan2025ai_pwrules_as_code}. Future work should pursue interactive validation loops (e.g., user feedback), richer semantic modeling, and expansive, evolving benchmarks to stress test real-world scenarios. Addressing these limitations will further strengthen fidelity, robustness, and auditability. Together, these advances position Policy$\rightarrow$Tests as a practical bridge from prose to enforcement, ready to be exercised against live policies and audited for real-world compliance.

\bibliography{aaai2026}

@inproceedings{cheng2025pii,
  author    = {Shuai Cheng and Shu Meng and Haitao Xu and others},
  title     = {Effective PII Extraction from LLMs through Augmented Few-Shot Learning},
  booktitle = {Proceedings of the 34th USENIX Security Symposium (USENIX Security 2025)},
  year      = {2025},
  address   = {Seattle, WA, USA},
  month     = {August},
  isbn      = {978-1-939133-52-6}
}

@inproceedings{shen2024jailbreak,
  author    = {Xinyue Shen and Zeyuan Chen and Michael Backes and others},
  title     = {{``Do Anything Now'': Characterizing and Evaluating In-The-Wild Jailbreak Prompts on Large Language Models}},
  booktitle = {Proceedings of the 2024 ACM SIGSAC Conference on Computer and Communications Security (CCS '24)},
  year      = {2024},
  address   = {Salt Lake City, UT, USA},
  month     = {October},
  pages     = {1671--1685}
}

@inproceedings{carlini2021extracting,
  author    = {Nicholas Carlini and Florian Tram{\`e}r and Eric Wallace and others},
  title     = {Extracting Training Data from Large Language Models},
  booktitle = {Proceedings of the 30th USENIX Security Symposium (USENIX Security 2021)},
  year      = {2021},
  month     = {August},
  pages     = {2633--2650},
}

@inproceedings{wei2024cotaeval,
  author    = {Boyi Wei and Weijia Shi and others},
  title     = {Evaluating Copyright Takedown Methods for Language Models},
  booktitle = {Advances in Neural Information Processing Systems 37 (NeurIPS 2024), Datasets and Benchmarks Track},
  year      = {{2024}},
}

@misc{EU_2024_AI_Act,
  author = {{European Union}},
  title  = {Regulation (EU) 2024/1689 of the European Parliament and of the Council of 13 June 2024 laying down harmonised rules on artificial intelligence (Artificial Intelligence Act)},
  year   = {2024},
  note   = {Official Journal of the European Union, L 2024/1689, 12 July 2024},
}

@article{NIST_2023_AI_RMF_1_0,
  title={Artificial intelligence risk management framework (AI RMF 1.0)},
  author={NIST,AI},
  pages={100--1},
  year={2023}
}

@misc{OECD_2019_AI_Principles,
  author = {{Organisation for Economic Co-operation and Development}},
  title  = {Recommendation of the Council on Artificial Intelligence},
  year   = {2019},
}

@misc{ISOIEC_42001_2023,
  author = {{International Organization for Standardization} and {International Electrotechnical Commission}},
  title  = {ISO/IEC 42001:2023 Information technology, Artificial intelligence and Management system},
  year   = {2023},
}

@misc{NIST_2023_Playbook,
  author = {{National Institute of Standards and Technology}},
  title  = {NIST AI RMF Playbook},
  year   = {2023},
}

@misc{PDPC2021AIVerifyMVP,
  author       = {Personal Data Protection Commission},
  title        = {Developing the MVP for AI Governance Testing Framework},
  year         = {2021},
  month        = {July},
  day          = {14},
  note         = {States the framework uses a mix of technical/statistical tests and process checks. Accessed 17 Oct 2025}
}

@inproceedings{ouyang2022instructgpt,
  title        = {Training Language Models to Follow Instructions with Human Feedback},
  author       = {Ouyang, Long and Wu, Jeff and Jiang, Xu and others},
  booktitle    = {Advances in Neural Information Processing Systems (NeurIPS)},
  year         = {2022},
}

@article{bai2022constitutional,
  title={Constitutional ai: Harmlessness from ai feedback},
  author={Bai, Yuntao and Kadavath, Saurav and others},
  journal={arXiv preprint arXiv:2212.08073},
  year={2022}
}

@manual{microsoft,
  author        = {{Microsoft Corporation}},
  title         = {Responsible AI Standard v2: General Requirements},
  organization  = {Microsoft Corporation},
  year          = {2022},
  note          = {Microsoft Responsible AI Standard, publicly released June 2022}
}

@article{cohen1960,
  author    = {Jacob Cohen},
  title     = {A Coefficient of Agreement for Nominal Scales},
  journal   = {Educational and Psychological Measurement},
  volume    = {20},
  number    = {1},
  pages     = {37--46},
  year      = {1960},
  publisher = {SAGE Publications},
  doi       = {10.1177/001316446002000104}
}

@techreport{kennan2025ai_pwrules_as_code,
  author       = {Ariel Kennan and Lisa Singh and Alessandra Garcia Guevara and Mohamed Ahmed and Jason Goodman},
  title        = {AI-Powered Rules as Code: Experiments with Public Benefits Policy},
  institution  = {Beeck Center for Social Impact + Innovation, Digital Government Hub, Georgetown University},
  year         = {2025},
  month        = {mar},
  note         = {Last updated September 4, 2025; documents experiments translating SNAP and Medicaid policy into machine‐readable rules.}
}

@book{krippendorff2018,
  author    = {Klaus Krippendorff},
  title     = {Content Analysis: An Introduction to Its Methodology},
  edition   = {4},
  publisher = {SAGE Publications},
  year      = {2018},
  address   = {Thousand Oaks, CA}
}

@regulation{hipaa,
  key          = {HIPAA 2003},
  title        = {Standards for Privacy of Individually Identifiable Health Information (HIPAA Privacy Rule), 45~C.F.R.~Part~164~Subpart~E},
  journal      = {Code of Federal Regulations},
  volume       = {45},
  part         = {164},
  subpart      = {E},
  year         = {2003},
  note         = {Privacy Rule under the Health Insurance Portability and Accountability Act (HIPAA)}
}

@manual{nemo_guardrails_overview,
  title        = {NeMo Guardrails Overview},
  author       = {{NVIDIA}},
  organization = {NVIDIA Corporation},
  year         = {2025},
  note         = {Accessed 2025-10-17}
}

@misc{guardrails_ai_github,
  title        = {Guardrails: A Framework for Building Reliable AI Applications},
  author       = {{Guardrails AI}},
  year         = {2025},
  note         = {Documentation accessed 2025-10-17}
}

@misc{openai_evals,
  title        = {Evals: A Framework for Evaluating LLMs and LLM Systems},
  author       = {{OpenAI}},
  year         = {2025},
  note         = {Repository accessed 2025-10-17}
}

@manual{promptfoo_docs,
  title        = {promptfoo Documentation: Intro and Getting Started},
  author       = {{promptfoo}},
  organization = {promptfoo},
  year         = {2025},
  note         = {Accessed 2025-10-17}
}

@inproceedings{palla2025policyasprompt,
  title        = {Policy-as-Prompt: Rethinking Content Moderation in the Age of Generative AI},
  author       = {Palla, Karthik and Li, Aijun and Hosseini, Hossein and Deng, Yue and Sandholm, Thomas et al.},
  booktitle    = {Proceedings of the ACM Web Conference 2025},
  year         = {2025},
  publisher    = {ACM},
}

@misc{openfisca_docs,
  title        = {What is OpenFisca},
  author       = {{OpenFisca contributors}},
  year         = {2025},
  note         = {Accessed 2025-10-17}
}

@misc{omg_dmn_overview,
  title        = {Precise specification of business decisions and business rules — Overview},
  author       = {{Object Management Group}},
  year         = {2025},
  note         = {Accessed 2025-10-17}
}

@misc{opa_docs_intro,
  title        = {Open Policy Agent: Introduction},
  author       = {{Open Policy Agent maintainers}},
  year         = {2025},
  note         = {Accessed 2025-10-17}
}

@misc{ibm_graphrag_tutorial,
  title        = {Implementing Graph RAG Using Knowledge Graphs},
  author       = {{IBM}},
  year         = {2024},
  note         = {Accessed 2025-10-17}
}

@article{complai2024,
  title        = {COMPL-AI Framework: A Technical Interpretation and LLM Benchmarking Suite for the EU Artificial Intelligence Act},
  author       = {Guldimann, Philipp and Spiridonov, Alexander and Staab, Robin and Jovanovi{\'c}, Nikola et al.},
  journal      = {arXiv preprint arXiv:2410.07959},
  year         = {2024},
}

@article{sgbench2024,
  title={Sg-bench: Evaluating llm safety generalization across diverse tasks and prompt types},
  author={Mou, Yutao and Zhang, Shikun and Ye, Wei},
  journal={Advances in Neural Information Processing Systems},
  volume={37},
  pages={123032--123054},
  year={2024},
}

@misc{toloka_taubench_policy_agents,
  title        = {TAU-Bench Extension: Benchmarking Policy-aware Agents in Realistic Settings},
  author       = {{Toloka}},
  year         = {2025},
  note         = {Accessed 2025-10-17}
}

\appendix

\section{Appendix}
\subsection{Policy$\rightarrow$Tests DSL (v1)}
\label{app:dsl}

\textbf{Overview.}
Each extracted rule is a JSON object with fixed fields for provenance, scope, hazard, conditions, exceptions, requirements, evidence, severity, and testability. We keep the DSL compact and provenance-first: (i) all rules carry \texttt{source.doc}, \texttt{source.citation}, and \texttt{source.span\_id}; (ii) scope uses simple string arrays to stay model-agnostic; (iii) \texttt{testability} records the rationale and which evidence channels are appropriate for verification. Below is the JSON Schema used in our experiments.

\begin{lstlisting}[language=json,frame=single,
  basicstyle=\ttfamily\tiny,
  breaklines=true,
  aboveskip=0.8\baselineskip,  % space before
  belowskip=0.8\baselineskip   % space after
]
{
  "$schema": "https://json-schema.org/draft/2020-12/schema",
  "title": "Policy -> Tests DSL (ultra-mini)",
  "type": "object",
  "additionalProperties": false,
  "required": ["rule_id","source","scope","requirement","is_testable","testability"],
  "properties": {
    "rule_id": {"type":"string"},
    "source": {
      "type":"object", "additionalProperties": false,
      "required":["doc","citation","span_id"],
      "properties": {
        "doc":{"type":"string"},
        "citation":{"type":"string"},
        "span_id":{"type":"string"}
      }
    },
    "scope": {
      "type":"object", "additionalProperties": false,
      "required":["actor"],
      "properties": {
        "actor":{"type":"array","items":{"type":"string"}},
        "data_domain":{"type":"array","items":{"type":"string"}},
        "context":{"type":"array","items":{"type":"string"}}
      }
    },
    "hazard": {"type":"string"},
    "requirement": {"type":"string"},
    "severity": {"type":"string","enum":["low","medium","high"]},
    "is_testable": {"type":"boolean"},
    "testability": {
      "type":"object", "additionalProperties": false,
      "required":["evidence_signals"],
      "properties": {
        "evidence_signals":{"type":"array","items":{"type":"string"}},
        "reason":{"type":"string"}
      }
    }
  }
}
\end{lstlisting}

\subsection{Operational cost and runtime}
We report indicative pipeline efficiency to contextualize automation benefits. Across four documents the pipeline processed 42{,}465{,}118 input tokens for about \$20 total and ran about 30 minutes to 3 hours per document, depending on document length, clause density, and model choice. For comparison, experienced annotators typically require 5--8 minutes per atomic rule, which yields tens of hours of manual work even before adjudication. Summary figures appear in Table~5.

\subsection{Assessing the safety impact of generated rules - A case study}

We evaluate whether rule enforcement reduces unsafe behavior while preserving robustness across prompt types by
comparing a baseline assistant with the same assistant instrumented with guardrails. Three HIPAA-derived, I/O-testable rules are encoded as NeMo output rails \cite{ nemo_guardrails_overview} and applied to an assistant. The rules target three
high risk themes in our corpus, namely HIPAA permitted
PHI use only, no genetic PHI for underwriting, and marketing payment must be disclosed. On 60 prompts (20 clean, 20 obfuscated, 20 compositional), an LLM judge evaluates each response against a JSON rule description and returns pass/fail. We measure violation rate overall and by bucket and
report robustness deltas relative to the clean bucket; the prompt set is released with our code for reproducibility, and Table 4 summarizes results.

\begin{table}[t]
\centering
\small
\begin{tabularx}{0.425\textwidth}{@{}l l c c@{}}
\toprule
\textbf{System} & \textbf{Bucket} & \textbf{Violation rate} & \textbf{$\Delta$ vs clean} \\
\midrule
Baseline       & Clean          & 0.02 & 0.00 \\
Baseline       & Obfuscated     & 0.58 & +0.56 \\
Baseline       & Compositional  & 0.42 & +0.40 \\
Baseline       & Overall        & 0.34 & ---  \\
\midrule
Guarded        & Clean          & 0.00 & 0.00 \\
Guarded        & Obfuscated     & 0.08 & +0.08 \\
Guarded        & Compositional  & 0.06 & +0.06 \\
Guarded        & Overall        & 0.05 & ---  \\
\bottomrule
\end{tabularx}
\caption{Violation rates with clean defined as benign prompts. $\Delta$ uses each system's clean rate as anchor. Overall is the mean across 60 prompts.}
\end{table}

\begin{table}[t]
\centering
\begingroup
\setlength{\tabcolsep}{3pt}
\renewcommand{\arraystretch}{0.95}
\scriptsize
\begin{tabularx}{0.47\textwidth}{@{}l r c r r@{}}
\toprule
\textbf{Doc} & \textbf{Rules} & \textbf{RT (h)} & \textbf{Pipe. cost (USD)} & \textbf{Ann. hrs (exp)} \\
\midrule
EU AI Act (Arts.\ 8--15)   & 51  & 0.6--1.2 & \$3.5 & 4.3--6.8 \\
NIST AI RMF Profiles       & 117 & 1.0--2.0 & \$6.0 & 9.8--15.6 \\
HIPAA Privacy Rule         & 77  & 0.5--1.0 & \$4.0 & 6.4--10.3 \\
Microsoft RAI Standard v2  & 140 & 1.0--3.0 & \$6.5 & 11.7--18.7 \\
\bottomrule
\end{tabularx}
\endgroup
\caption{Compact runtime and pipeline cost summary. Abbrev.: RT = pipeline runtime; \emph{Pipe. cost (USD)} = pipeline cost in USD; \emph{Ann. hrs (exp)} = manual annotation hours for experienced raters (5--8 minutes per rule).}
\label{tab:runtime_cost_compact2}
\end{table}

\subsection{Ambiguity and failure modes}
Ambiguous or context-dependent clauses remain challenging. We mitigate ambiguity with paragraph-level chunking to preserve local qualifiers, clause mining to surface deontic cues, a rubric-driven judge with minimal repair to catch missing hazard or scope, and counterfactual probing to expose polarity sensitivity. These controls reduce errors but do not eliminate them. Recurrent failures include softened or dropped qualifiers, scope misassignment when key cues are nonlocal or cross-referenced, and unclear temporal or conditional language. SMT checks, evidence gating, and de-duplication target contradictions, verifiability, and redundancy rather than ambiguity. Additional failure modes include nested exceptions and negations, overlapping or multi-party scopes, and implicit conditions that fall outside the extracted span. The pipeline can also over-normalize nuanced qualifiers into coarse schema slots, blurring distinctions between “may,” “should,” and “shall.” Document-specific drafting quirks and domain shifts degrade clause-mining and extraction heuristics, leading to uneven performance across documents and domains.
\end{document}